\documentclass{article}

\usepackage{microtype}
\usepackage{graphicx}
\usepackage{subfigure}
\usepackage{booktabs} 

\usepackage{amsmath}
\usepackage{algorithm}
\usepackage[noend]{algorithmic}
\usepackage{hyperref}

\usepackage[accepted]{icml2021}

\usepackage{enumitem}
\usepackage{natbib}

\icmltitlerunning{Simeon -- Secure Federated Machine Learning}

\begin{document}

\twocolumn[
\icmltitle{Simeon -- Secure Federated Machine Learning Through Iterative Filtering}




\begin{icmlauthorlist}
\icmlauthor{Nicholas Malecki}{to}
\icmlauthor{Hye-young Paik}{to}
\icmlauthor{Aleksandar Ignjatovic}{to}
\icmlauthor{Alan Blair}{to}
\icmlauthor{Elisa Bertino}{goo}
\end{icmlauthorlist}

\icmlaffiliation{to}{School of Computer Science and Engineering, University of New South Wales, Sydney, Australia}
\icmlaffiliation{goo}{Department of Computer Science, Purdue University, West Lafayette, Indiana, United States}

\icmlcorrespondingauthor{Nicholas Malecki}{n.malecki@unsw.edu.au}
\icmlcorrespondingauthor{Hye-young Paik}{h.paik@unsw.edu.au}

\icmlkeywords{Federated Machine Learning, Adversarial Attacks, Iterative Filtering}

\vskip 0.3in
]



\printAffiliationsAndNotice{}  

\begin{abstract}
Federated learning enables a global machine learning model to be trained collaboratively by distributed, mutually non-trusting learning agents who desire to maintain the privacy of their training data and their hardware. 
A global model is distributed to clients, who perform training, and submit their newly-trained model to be aggregated into a superior model. However, federated learning systems are vulnerable to interference from malicious learning agents who may desire to prevent training or induce targeted misclassification in the resulting global model. A class of Byzantine-tolerant aggregation algorithms has emerged, offering varying degrees of robustness against these attacks, often with the caveat that the number of attackers is bounded by some quantity known prior to training. This paper presents Simeon: a novel approach to aggregation that applies a reputation-based iterative filtering technique to achieve robustness even in the presence of attackers who can exhibit arbitrary behaviour. We compare Simeon to state-of-the-art aggregation techniques and find that Simeon achieves comparable or superior robustness to a variety of attacks. Notably, we show that Simeon is tolerant to sybil attacks, where other algorithms are not, presenting a key advantage of our approach.
\end{abstract}

\section{Introduction}
\label{sec:intro}

\emph{Federated learning} is a technique for training a machine learning model by distributing training across disparate independent learning agents, each with their own private training data  \cite{mcmahan2017}. At each training iteration, each agent receives a copy of the current \emph{global model}, performs training using its own data, and submits its own trained model to be aggregated into a new global model  \cite{mcmahan2017}. Federated learning systems are characterised by clients' autonomy to freely associate and dissociate themselves from the system, the maintenance of the privacy of training data, and clients' capability to control how much of their data is shared with the system  \cite{li2019}. This technique has demonstrable, practical viability in distributed predictive text  \cite{hard2018, beaufays2019}, and for training via healthcare data   \cite{grama2020robust}, and other problem domains where the privacy of training data is paramount.

Federated learning systems can have different configurations   \cite{li2019}.
For the purposes of this paper, we focus on federated learning systems that are horizontal (each clients’ data sets share the same feature space, but different sample space), public (a large number of voluntarily participating clients, each with small data sets), and centralised (model aggregation is performed at a centralised location), as these systems have received the most research attention and are the most readily applicable for real-world applications.

Federated learning’s guarantees of client autonomy and privacy create significant security challenges. Operators of these systems can rarely control the entry and exit of participants, creating the opportunity for the entry of malicious clients who may desire to influence or prevent training. Worse, to preserve privacy, clients and their data are considered opaquely, precluding client inspection or profiling, ultimately rendering the detection and exclusion of malicious clients challenging. These difficulties are compounded by the tendency for learning agents to train on highly disparate data sets of varying sizes, presenting an obstacle to separating truly anomalous behaviour from natural variation between clients and possibly offering an avenue for an attack   \cite{baruch2019}.

Broadly speaking, the literature divides attacks on federated learning into \emph{data poisoning} and \emph{model poisoning}  \cite{kairouz2019advances}. \emph{Data poisoning} uses compromised training examples to induce some attacker-desired behaviour  \cite{lyu2020threats}.
It has been demonstrated that federated learning systems are generally resilient towards dirty-label data poisoning, as the model aggregation process attenuates the differences between the poisoned models and the benign models  \cite{fang2019}. 

Meanwhile, \emph{model poisoning} involves using compromised learning agents to submit manipulated models for aggregation, and may be further categorised into \emph{targeted} model poisoning (using model poisoning to induce an attacker-chosen misclassification), and \emph{untargeted} model poisoning (using model poisoning to prevent training from converging, or to induce training to converge to an ineffective model)  \cite{bhagoji2019}. The literature shows that model poisoning attacks are effective against current state-of-the-art aggregation techniques in federated learning systems \cite{fang2019, bhagoji2019, bagdasaryan2019}. The capability for clients to enter and leave the system as they wish also leaves federated learning systems vulnerable to sybil attacks  \cite{fung2019} -- the rapid influx of malicious clients designed to induce some attacker-desired behaviour.

This paper explores a novel algorithm termed \emph{Simeon}, a Byzantine-robust aggregation mechanism against \textit{both} targeted and untargeted model poisoning attacks in federated learning systems. Simeon applies the concept of iterative filtering to permit secure aggregation, without requiring prior knowledge of an upper bound on the number of attackers. 
%
%
We demonstrate the robustness of Simeon against a range of model poisoning attacks, and benchmark the results with some of the cutting-edge aggregation algorithms that are designed to withstand Byzantine clients. 

\section{Related Work}






Most of the current aggregation algorithms in federated learning are vulnerable to model poisoning attacks \cite{kairouz2019advances,fang2019}. Particularly, aggregation mechanisms based on some linear combination of the submitted models have been shown to be trivially defeated by a single Byzantine peer who submits a model scaled according of the inverse of the linear combination used for aggregation. 

Krum was proposed as an alternative to such algorithms, and is not based on using a linear combination of submitted models \cite{blanchard2017}. Despite this, Krum has been demonstrated to be vulnerable to targeted \cite{bhagoji2019, elmhamdi2018, bagdasaryan2019} and untargeted \cite{fang2019} model poisoning attacks. Worse still, attacks have been found that are transferable between different aggregation techniques, indicating that federated learning is generally vulnerable to model poisoning techniques \cite{fang2019}. 

In particular, Bagdasaryan et~al.~\cite{bagdasaryan2019} demonstrated a targeted model poisoning attack on Krum which implemented a \emph{semantic backdoor}: targeted misclassification, triggered by some naturally-occurring (but attacker-chosen) feature of the input data, such as the presence of a particular pattern or object in an image. This was achieved by providing Byzantine clients with mislabelled data items containing the trigger characteristic, then applying scaling to the backdoor-trained model to overcome the attenuating effects of aggregation.

Existing defences against backdoor attacks like this require either a full control of the training process or an examination of training data which is not possible in the federated learning setting \cite{kairouz2019advances}.
Further, few of the latest aggregation techniques have proven robustness in the presence of sybil attacks. Krum and Bulyan require a priori knowledge of the upper bound on the number of attackers to guarantee resilience \cite{blanchard2017, elmhamdi2018}, whilst Multi-Krum (a Krum derivative) has been demonstrated to be vulnerable to sybil attacks  \cite{fung2019}. 

Simeon provides a successful defence against sybil attacks even when the system is attacked with an overwhelming number. We also show  Simeon's resistance to a semantic backdoor attack which is a state-of-the-art targeted model attack.


\section{Preliminaries and Problem Definition}
\label{sec:pre}
\subsection{Glossary}
\begin{itemize}[leftmargin=*]
    \setlength\itemsep{0.1cm}
        \item $n$: the number of clients.
    \item \emph{Data poisoning}: a method of attack involving compromised training examples to induce some attacker-desired behaviour.
    \item \emph{Model poisoning}: a method of attack involving compromised learning agents who can submit arbitrary models to induce some attacker-desired behaviour.
    \item \emph{Sybil attack}: a model-poisoning attack involving the addition of a large number of compromised clients to induce some attacker-desired behaviour.

    \item \emph{Aggregated model}: the result of the aggregation algorithm when applied to the models submitted by each client; namely, $F(M_s^0, ..., M_s^{n-1})$.
    \item $\eta$: the global learning rate; the proportion of the aggregated model that is applied~to~$M_g^{i-1}$ to calculate~$M_g^i$.
    \item $M_g^i$: the global model weights represented as a matrix, at the $i$th training round.
    \item \emph{Training round}: one round of:
        \begin{enumerate}
            \item Transmitting $M_{g}^{i-1}$ to all learning participants
            \item Performing a number of local epochs of training
            \item Transmitting $M_{s}$ to the central aggregation server
            \item Aggregating the models using the aggregation function into $M_g^i = (1 - \eta)M_g^{i-1} + \eta F(M_s^0, ..., M_s^{n-1})$
        \end{enumerate}
    \item $f$: the upper bound on the number of Byzantine clients.
    \item $\epsilon$: the degree of precision provided to Simeon; once the root mean square error between the reputation from the last iteration and the current iteration is $< 
    \epsilon$, the algorithm halts.
    \item $M_{bk}^i$: the backdoor-injected model trained by client $k$, with weights represented as a matrix, at the $i$th training round.

    \item $M_{sk}^i$: the model weights submitted by client $k$ represented as a matrix, at the $i$th training round.
    \item $\gamma$: the backdoor scaling factor.
    \item $c$: the number of replacements per batch during backdoor injection.
\end{itemize}

\subsection{Threat model and assumptions}

For the purposes of our investigation, we assume that all data used by clients for training is unavailable for inspection by the aggregation server, in line with the consideration for privacy that underscores federated learning \cite{mcmahan2017, li2019}. We also assume that the aggregation server is not malicious, and that its behaviour cannot be controlled by the attacker.

With respect to individual attackers, we assume a Byzantine threat model; that is, clients can exhibit arbitrary behaviour --- including seemingly-correct behaviour. This is the most broad category of means to induce failure because clients can do anything to induce failure in the system \cite{lamport1982}.  We then investigate the following five attacks:

\paragraph{{\underline{Noisy Clients Attack}:}} 
In this attack, Byzantine clients perform local training identically to benign clients before adding Gaussian noise with $\sigma = 1, \mu = 0$ to each weight in the submitted model. We use this attack as a baseline point of comparison to other attacks. 

\paragraph{\underline{Collusion Attack}:} In this attack,
100 weights are chosen and agreed-upon by all Byzantine clients prior to training. The chosen weights are then perturbed by amounts drawn from a Gaussian distribution with $\sigma = 1, \mu = 0$, also agreed-upon before training. We intend this attack as a more subtle evolution of the previous attack.

\paragraph{\underline{Backdoor Attack}:} 
This is a targeted model poisoning attack based on that described by Bagdasaryan et al. --- to this end, we make the same assumptions about the attacking clients, namely that attackers can:
\begin{itemize}[leftmargin=*]
    \item Control the local training of any compromised client.
    \item Control the hyperparameters and local training procedure of any compromised client.
    \item Modify the weights of compromised clients after training.
    \item Adaptively change training procedure for compromised clients from round to round  \cite{bagdasaryan2019}.
\end{itemize}

\paragraph{\underline{Sybil Attack}:} We strengthen the \textit{backdoor attack} by adapting it to a sybil attack.  We inject extra clients some time into the training process to enhance the attack's effectiveness. To this end, we impose the additional assumption that clients can join and leave the system arbitrarily between training rounds, and that there is no upper limit on the number of clients that can join the system.

\paragraph{\underline{Increasing Scaling Attack}:} In this attack, the colluding clients act benign at the start, then become increasingly malicious over time. To this end, the compromised clients send honest model updates at the start, and slowly increase the scaling parameter over time to boost the backdoor effect.

\section{Simeon - Byzantine robust aggregation}
\label{sec:simeon}

Having described a suite of attacks that we believe form a suitable benchmark for comparing aggregation algorithms, we now describe the details of our approach. \emph{Simeon}\footnote{Simeon I of Bulgaria was a ninth-century Bulgarian king known for his successful campaigns against the Byzantines. During his reign Bulgaria had its greatest territorial expansion ever, significantly larger than during the rule of Krum about 100 years earlier.} is a novel aggregation technique that applies iterative filtering to achieve robustness against the attack models described previously. Unique characteristics of Simeon include the use of a precision parameter to control the run-time of the algorithm, and the novelty of applying iterative filtering techniques to federated learning.

We largely base our approach on work by Rezvani et al. on the application of iterative filtering to an algorithm for aggregating readings from wireless sensor networks in the presence of malicious or compromised sensors  \cite{rezvani2014}
At a high level, Simeon is a direct adaptation of this algorithm to federated learning, by considering sensors as learning agents and sensor readings as model weights.

\subsection{Iterative filtering in brief}
To explain the main idea of iterative filtering, first introduced by Laureti et al.  \cite{laureti}, we start with an example. Assume that we have $n$ agents (such as sensors, for example) with each sensor $i$ ($1\leq i\leq n$) providing an evaluation $E(i,j)$ of $m$ quantities $q_j$ ($1\leq j\leq m$) and assume that the variance $v_i$ of each agent $i$ is known. One can then prove that, if the errors are Gaussians, the minimal variance unbiased estimation $\widehat{q}_j$ of each quantity $q_j$ is obtained as the weighted sum
\begin{align}\label{MLE}\widehat{q}_j=\sum_{i=1}^n\frac{1/v_i}{\sum_{j=1}^n1/v_j}E(i,j)\end{align}
and, in fact, such an estimator reaches the Cramer-Rao lower bound for the minimal possible variance of unbiased estimators. However, in practice, the variances of agents are unknown. Iterative filtering algorithms attempt to circumvent such a problem by iteratively simultaneously estimating the quantities $q_i$ of interest and the reliabilities $c_j$ of the agents. In the original algorithm from  \cite{laureti} the reliability $c_j$ of each agent $j$ was measured as the normalised reciprocal of its estimated variance, $c_j=\frac{1/\widehat{v}_j}{\sum_{i=1}^k1/\widehat{v}_i}$. The simple mean is taken as the initial estimate of the true values of the quantities $q_i$ measured and  the sample variance $\widehat{v}$ of each agent is then taken as an estimate of the true variance of each agent. In the subsequent round of estimation of the true values of the magnitudes measured, the simple mean is replaced by estimation given by \eqref{MLE} with $\widehat{v}_j$ in place of the true $v_j$. This, in turn, allows more precise estimation of the variance $\widehat{v}_j$ of each sensor $j$ and the whole process is repeated until the differences of the values of the estimated quantities in two consecutive rounds drop below a predetermined threshold. 
Such an iterative procedure often produces estimates extremely close to the maximum likelihood estimates given by \eqref{MLE} with \emph{true variances} rather than estimated ones; upon convergence such estimated variances are indeed often extremely close to the true, unknown variances. However, the reciprocal function has a pole at zero and, consequently, the estimates of each agent act as \emph{attractors} during the iterative procedure, and often the iterative procedure converges to the values of a single agent, assigning a negligible weight to the measurements of all other agents. There were several attempts to address this problem, with different degrees of success, see e.g.,  \cite{dekerchove} for example, where the reciprocal of the variances were replaced with an affine measure of the reliability of agents which does not suffer from attractors but which significantly increases the variance of the thus obtained estimator. In this paper we use an improvement of the iterative filtering algorithm first introduced in  \cite{rezvani2014} which we now describe.

\subsection{Algorithm description}
\begin{algorithm}[ht!]
\begin{algorithmic}
\caption{Simeon, aggregating on training round $k$}
\small
\REQUIRE $k >= 0, M = M_s^{0} ... M_s^{n-1}$
\STATE $t \leftarrow 0$
\LOOP
    \IF{$k = 0$}
        \STATE $E_k \leftarrow \frac{1}{n}\sum_{t=0}^{n-1}{M_s^t}$\\[3pt]
        \FOR{$i \in [0 .. n-1]$}
            \STATE $v_i \leftarrow \frac{1}{n}\sum_{j=1}^n{\rm MSE}(M_j, E)$
        \ENDFOR
        \FOR{$i \in [0 .. n-1]$}
            \STATE $c_i \leftarrow \sqrt[n]{\prod_{j=0}^{n-1}{\cfrac{e^{-{\rm MSE}(M_i, E_k)/2v_j}}{\sqrt{2\pi v_j}}}}$
        \ENDFOR
    \ELSE
        \STATE $E_k \leftarrow M \cdot W$
        \FOR{$i \in [0 .. n-1]$}
            \STATE $v_i \leftarrow {\rm MSE}(M_i, E)$
        \ENDFOR
        \FOR{$i \in [0 .. n-1]$}
            \STATE $c_i \leftarrow \sqrt[n]{\prod_{j=0}^{n-1}{\cfrac{e^{-v_i/2v_j}}{\sqrt{2\pi v_j}}}}$
        \ENDFOR
    \ENDIF
    \STATE $s \leftarrow \sum_{i=0}^{n-1}{c_i}$
        \FOR{$i \in [0 .. n-1]$}
            \STATE $w_i \leftarrow \cfrac{c_i}{s}$
        \ENDFOR
        \STATE $W = (w_0,\ldots,w_{n-1})$
    \IF{${\rm RMSE}(E, E_k) < \epsilon$}
        \STATE {\bf break}
    \ENDIF
    \STATE $t \leftarrow t + 1$
\ENDLOOP
\STATE $R\leftarrow(1/v_0,\ldots,1/v_{n-1})$
\STATE $s\leftarrow {\rm sum}(R)$
\STATE $R=R/s$; $E_k=M.R$ \\
\STATE {\bf return} $E_k$
\end{algorithmic}
\end{algorithm}
After each client $C = C_0 ... C_{n-1}$ has submitted their respective models $M = M_s^{0} ... M_s^{n-1}$ to the aggregation server at the end of round $k$:
\begin{enumerate}[leftmargin=*]
    \item At the first round of iteration ($t=0$) of the first round of learning ($k=0$) an initial estimate model $E_{0}$ is formed as the simple mean of $M_s^{0} ... M_s^{n-1}$. Also, an initial variance estimate, equal for all models is formed, equal to the mean sample variances of all models, ($v_i = \frac{1}{n-1}\sum_{j=0}^{n-1}{{\rm MSE}(M_j, E_{k-1})}, i \in [0, n-1]$). This is done to prevent attacks of type described in  \cite{rezvani2014}, where one of the colluding agents tries to match the mean of the measurements of the rest of the agents, skewed by the remaining colluding agents. Then, an initial credibility score $c_i$ for each model is then calculated as \[c_i = \sqrt[n]{\prod_{j=0}^{n-1}{\frac{e^{\frac{-{\rm MSE}(M_i, E_k)}{2v_j}}}{\sqrt{2\pi v_j}}}}\]
    
    
    
    \item The next iteration begins: the new variance $v_i$ of each model $M_i$  is estimated as the mean square error between $M_i$ and the estimate $E_{k-1}$ ($v_i = {\rm MSE}(M_i, E_{k-1}), i \in [0, n-1]$).
    \item A credibility score $c_i$ for each model is then calculated as 
    \[c_i = \sqrt[n]{\prod_{j=0}^{n-1}{\frac{e^{\frac{-v_i}{2v_j}}}{\sqrt{2\pi v_j}}}}= \sqrt[n]{\prod_{j=0}^{n-1}{\frac{e^{\frac{-\mathrm{sum}(M_i-E)^2}{2v_j}}}{\sqrt{2\pi v_j}}}}.\]
    To understand this we look at the second, slightly expanded form. The credibility of a model is the geometric mean of the likelihoods of the values in that model, estimated from the perspective of all of the models. It is easy to see that such a likelihood will be high for models which are close to a sufficiently large number of other models. 
    \item The new estimate of the aggregate values is calculated as the sum of $M_s^{j} \forall j \in [0, n-1]$, weighted according to their corresponding credibility scores $c_j$, normalised by the sum of all $c_i$'s.
    \item The algorithm halts if the root mean square error between the estimates in two consecutive rounds of iteration is less than $\epsilon$. Otherwise, the algorithm repeats from (1). 
    
    \item The reason why during the iterations we use the credibility of models rather than the reciprocals of variances is that $1/v_i$ converges to infinity as $v_i$ gets small. However, while $c_i$ grows for a while as variance decreases, as it gets extremely close to 0 $c_i$ actually starts converging to 0 and so the models no longer act as attractors, making the algorithm extremely robust; see the Figure \ref{fig:plot}.
    
    \begin{figure}[!h]
    \centering
     \includegraphics[width=0.6\columnwidth]{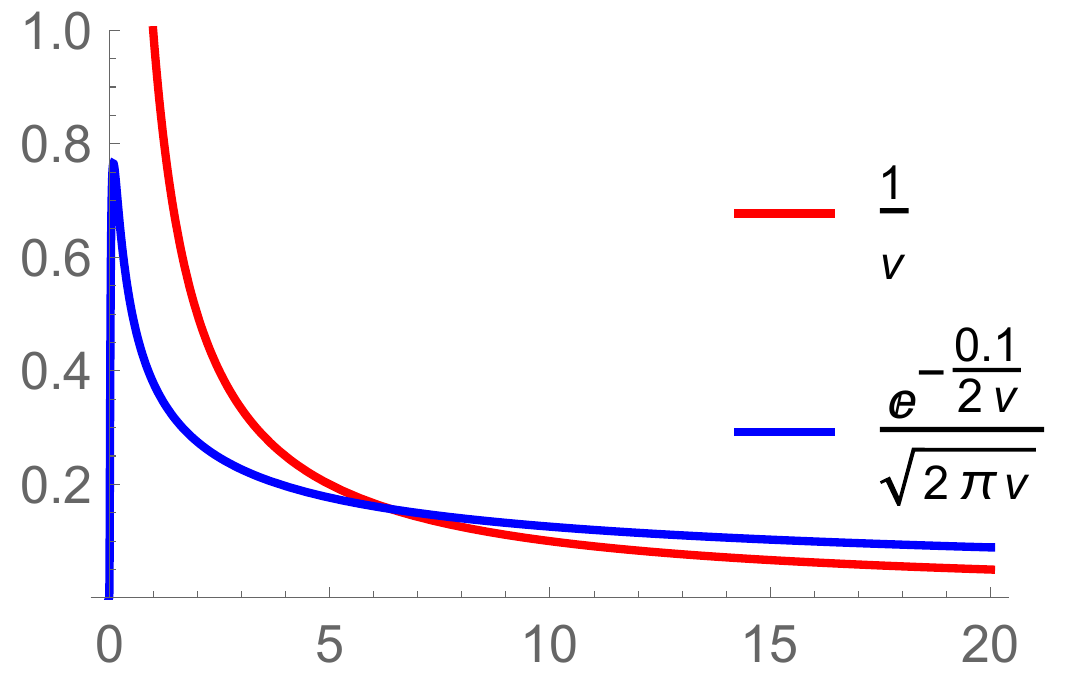}

    \caption{$c_i$ vs. $1/v_i$}
    \label{fig:plot}
    \end{figure}

    \item The final estimate $E_k$ of round $k$ is calculated as the sum of $M_s^{j} \forall j \in [0, n-1]$, weighted by the reciprocals of the final variances of the models, normalised by the sum of all such reciprocals. Thus, the credibilities were used only to estimate variances of models in a robust way; the final estimate is still obtained by using \eqref{MLE} with thus estimated variances.
\end{enumerate}


\subsection{Benchmarking Simeon}

A key advantage of our approach is that iterative filtering enables us to perform aggregation without a priori knowledge of the upper bound on the number of Byzantine clients. In fact, if the malicious clients provide significantly different models from the majority consensus, during the course of iteration they will be assigned very small weights  and thus will be marginalised. Thus, unlike with Krum, malicious clients are effectively excluded without any hard threshold on their number. If the malicious clients provide significantly different models between themselves, they will be marginalised even if they are in majority. The only way to defeat our aggregation system is if the colluding agents are in majority and all provide very similar skewed models; however, such behaviour makes the colluding agents detectable by the conventional means which estimate the divergence of the newly proposed models from the previous aggregated model. 

Our technique also provides robustness against sybil attacks, as new clients who submit models different from the 'consensus' will have a high variance and thus, a low credibility, leading to their exclusion from aggregation.


As points of comparison against Simeon, we investigated Krum \cite{blanchard2017}, Bulyan with Krum as the selection function \cite{elmhamdi2018}, and coordinate-wise median \cite{yin2018}. We also compared Simeon against federated averaging as a baseline \cite{mcmahan2017}. Krum involves taking the pairwise Euclidean distance between each submitted model, then selecting the $n - f - 2$ closest models to each. Models are then scored based on the sum of these distances. The model with the highest score is the output of Krum \cite{blanchard2017}. Bulyan is a meta-aggregation technique which uses a selection function to choose (without replacement) a subset of submitted models to consider for aggregation. The mean for each weight amongst the selected subset is then used as the value for the corresponding weight in the aggregated model \cite{elmhamdi2018}. For our investigation we chose Krum as the selection function because past research has focused on the case where Bulyan is used with Krum \cite{fang2019, elmhamdi2018, baruch2019}, although in theory other sampling functions can be used. Coordinate-wise median involves using the median of each coordinate amongst the submitted weights as the corresponding coordinate in the aggregated model \cite{yin2018}. Federated averaging involves taking a linear combination of each submitted model, typically weighted by the amount of test data reported to be available to the client \cite{mcmahan2017}. This algorithm is not Byzantine-robust, but serves as a baseline comparison for our investigation.

Our experiments were run on four nodes, each running an
Intel Xeon Gold CPU with an Nvidia Tesla V100-SXM2
32GB, on a 64-bit version of CentOS.

\section{Noisy clients}
\label{sec:noisy}

\paragraph{Experiment implementation and design}
As a baseline, we performed three analyses involving 20 clients, of which 2, 4, then 6 exhibit Byzantine behaviour, for an overall Byzantine client ratio of 10\%, 20\% and 30\% respectively. Our experiment involved a CIFAR-10 image classification model consisting of three VGG blocks, with a convolutional layer with 32, 64, and 128 output channels respectively. Batch normalisation and dropout were applied to each VGG block. We trained using a batch size of 64, using sparse categorical cross entropy as the loss function. 50,000 of the total 60,000 data items were reserved for training, with the remaining 10,000 reserved for validation. At the start of each round of training, each client received the global model and performed one epoch of training using stochastic gradient descent with a learning rate of 0.01 and a momentum value of 0.9 on an equally-sized, disjoint shard of the CIFAR-10 data set. For the purposes of this experiment, Byzantine clients performed the same number of training epochs as non-Byzantine clients, and added Gaussian noise to each weight, with $\mu = 0$ and $\sigma = 1$, before submitting the mutated model for aggregation. The aggregation function is then applied using all of the submitted models to produce an updated global model, implying a global learning rate of 1.0 (the aggregated model entirely replaces old global model).

We compared the performance of Simeon against three other Byzantine-robust aggregation methods: coordinate-wise median, Krum, and Bulyan (with Krum as the selection function). We also compared the performance of our algorithm against federated averaging. To measure the effectiveness of each algorithm, we measured the sparse categorical accuracy of the global model against the 10,000 validation items after each round of training and aggregation. The experiment was run for 250 rounds of training before being halted. For Krum and Bulyan, we provided an accurate upper bound for the number of Byzantine clients by setting $f$ to the actual number of Byzantine clients (or as close as possible for Bulyan\footnote{Bulyan requires $f \geq \cfrac{n - 3}{4}$, and thus for the 30\% test case we set $f=4$ to ensure it performed optimally}). Simeon was configured with $\epsilon = 10^{-7}$. We also tested the performance of Krum and Bulyan when given an inaccurate upper bound by setting $f = 0$ for Krum and $f = 2$ for Bulyan in the 20\% and 30\% Byzantine test cases.

\begin{figure*}
        \begin{center}
        \includegraphics[width=.316\linewidth]{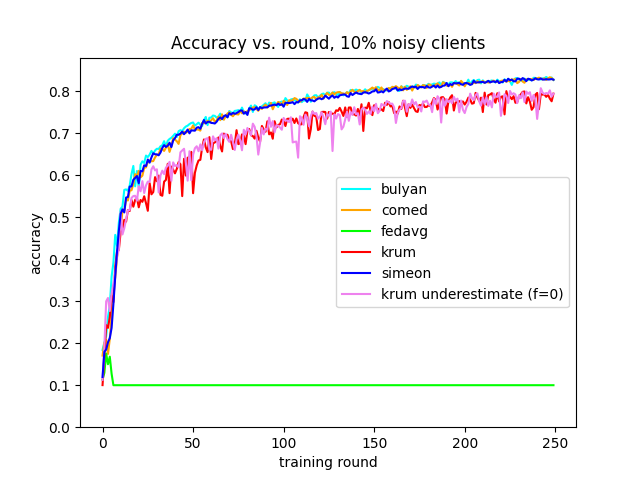}
        \quad \includegraphics[width=.316\linewidth]{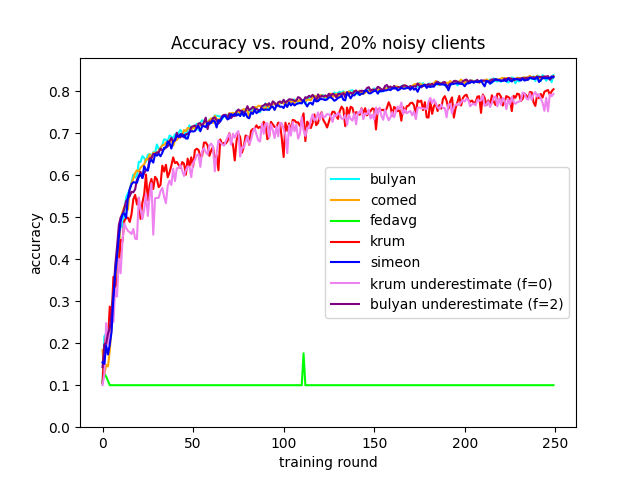}
        \quad \includegraphics[width=.316\linewidth]{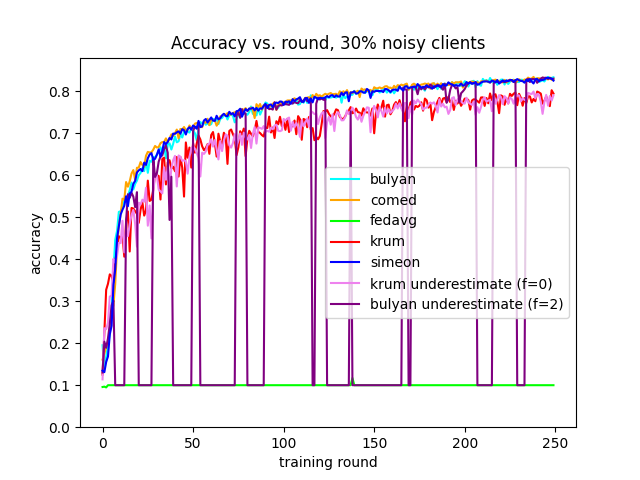}
        \end{center}
        \vspace{-1.2em}
        \caption[Noisy clients]{Left to Right: accuracy achieved with each aggregation algorithm in the presence of 10\%, 20\%, and 30\% noisy clients.}
        \label{fig:noisy}
    \end{figure*}

\paragraph{Results and analysis}
As shown in Figure~\ref{fig:noisy}, federated averaging fails to train in all three cases, collapsing to a rate of accuracy comparable to a random classifier. This is because the sum of the noise values from the Byzantine clients effectively increases the standard deviation of the noise applied to the global model. Hence, the global model is overwhelmingly likely to be totally distorted by noise when using federated averaging as the aggregation function. This is not unexpected: federated averaging is not designed for Byzantine-robustness  \cite{mcmahan2017}.

We note that the other algorithms perform well across the board in all three cases. Krum excludes noisy clients because they will generally not be clustered around any particular weight vector (i.e. their Euclidean distance to other models will be high). Because the noise added to each Byzantine client differs between clients, the noisy models will not even be clustered around each other, and thus they are trivially excluded by Krum from selection. Coordinate-wise median would naturally output models close to the value of the non-Byzantine clients’ models because the clients are overwhelmingly non-Byzantine. By taking the median, each coordinate will generally be close to the coordinate values submitted by benign clients, and hence, coordinate-wise median is robust to this attack. Meanwhile, Bulyan involves excluding the clients using some other sampling mechanism (Krum in our case). For this reason, the same argument for Krum's robustness applies to Bulyan using Krum as the selection function. However, we do note degraded performance for Bulyan when the upper bound on the number of Byzantine clients is not respected -- this has been noted as a key weakness of Bulyan and Krum \cite{fung2019}.

Simeon performs well in this example: the noise applied to the model means that the variance between the noisy model and the other models will be high. Hence, the credibility score granted to the noisy models will be very low, and they will not be weighted strongly when performing aggregation. In fact, we found that Simeon usually entirely excluded the noisy clients from aggregation. From these results, we concluded that Simeon is greatly resilient to this attack, trivial as it is.

\section{Collusion attacks}
\label{sec:collusion}

\paragraph{Experiment implementation and design}
Our second experiment used the same CIFAR-10 classification model described above, with the same ratios of Byzantine clients (10\%, 20\%, and 30\%). Accuracy and loss were calculated in the same manner. As before, at the start of each round of training, each client receives the global model and performs one epoch of training using stochastic gradient descent on the complete CIFAR-10 data set with a learning rate of 0.01 and a momentum value of 0.9. For the purposes of this experiment, Byzantine clients performed the same number of epochs of training as non-Byzantine clients, and added Gaussian noise with $\mu = 0$ and $\sigma = 1$ to a pre-determined subset of weights, agreed upon by all Byzantine clients to simulate an untargeted model poisoning attack based on a collusion strategy. This Gaussian noise was also agreed upon by all Byzantine clients prior to training, such that each Byzantine client applies the same noise to the same subset of weights. As with the previous experiment, the aggregation function is then applied using all of the submitted models to produce the new global model (an implied global learning rate of 1.0).

For this analysis, we compared Simeon against the same aggregation methods as before: coordinate-wise median, Krum, Bulyan (with Krum as the selection function), and federated averaging. We measured the sparse categorical accuracy of the global model against the 10,000 validation items after each round of training and aggregation, also running the experiment run for 250 rounds of training. As before, Krum and Bulyan were provided an accurate upper bound for the number of Byzantine clients by setting $f$ to the actual number of Byzantine clients (or as close as possible for Bulyan). Simeon was configured with $\epsilon = 10^{-7}$. Again, we also tested the performance of Krum and Bulyan when given an inaccurate upper bound by setting $f = 0$ for Krum and $f = 2$ for Bulyan in the 20\% and 30\% Byzantine test cases.

\paragraph{Results and analysis}

    \begin{figure*}
        \begin{center}
        \includegraphics[width=.33\linewidth]{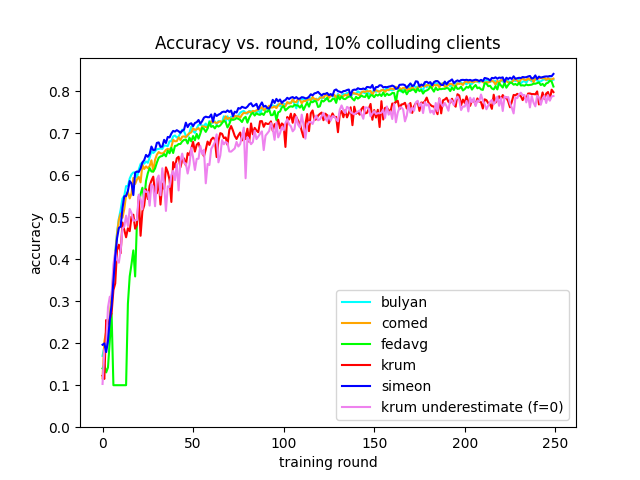}
        \includegraphics[width=.33\linewidth]{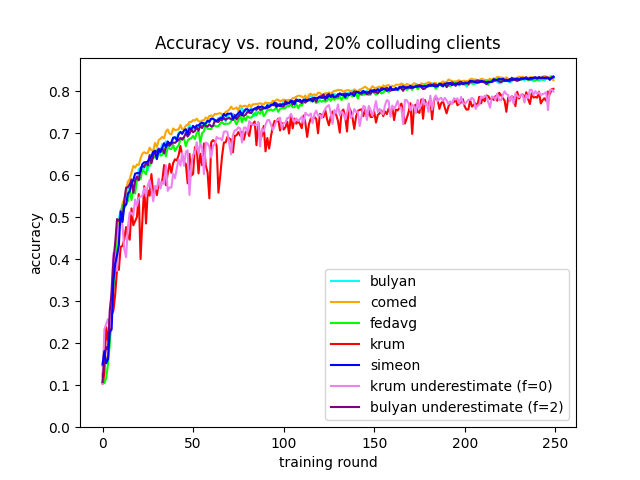}
        \includegraphics[width=.33\linewidth]{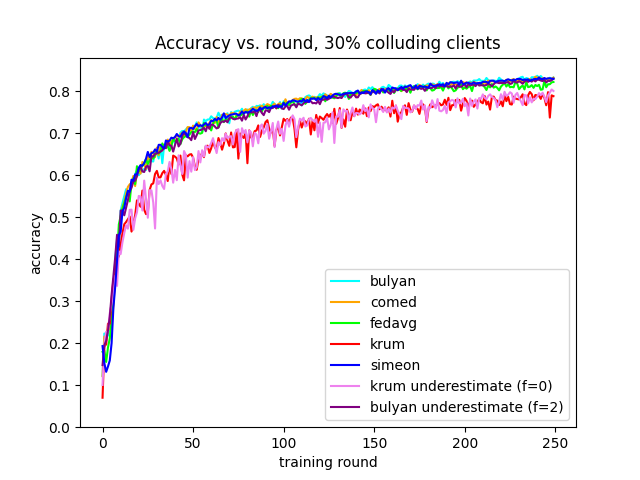}
        \end{center}
        \caption[Colluding clients]{Left to Right: accuracy achieved with each aggregation algorithm in the presence of 10\%, 20\%, and 30\% colluding clients.}
        \label{fig:colluding}
    \end{figure*}

The results are shown in Figure~\ref{fig:colluding}. We found that all of the algorithms were resilient to this attack in all three test cases -- federated averaging included. This is likely because the small number of perturbations resulted in a negligible impact on the final model. Our analysis of the results for the other algorithms follows a similar trend as the noisy clients case. Coordinate-wise median remains robust to this attack because the clients are overwhelmingly non-Byzantine, and thus the median coordinates will generally be close to the coordinate values submitted by non-Byzantine models. As before, Bulyan mitigates the effect of the noise by taking the trimmed mean after performing Krum sampling \cite{elmhamdi2018}, which should be unaffected by the noise. Notably,  Krum is still resilient to this attack: despite the fact that the colluding models are relatively clustered with each other, they are not numerous enough for Krum to select them, as the more numerous non-Byzantine models will be more clustered amongst each other.



\section{Backdoor attacks}
\label{sec:backdoor}

\paragraph{Experiment implementation and design}
We performed an experiment involving a CIFAR-10 classification model consisting of three VGG blocks, with a convolutional layer with 32, 64, and 128 output channels respectively. Batch normalisation and dropout were applied to each VGG block. On a control run with no Byzantine clients and 20 benign clients using federated averaging, this network architecture achieved a rate of accuracy of 85\% and a rate of misclassification of 9.6\% after 250 rounds of training. Each experiment run involved 20 clients, and the number of Byzantine clients varied between 10\%, 20\%, and 30\%.

Byzantine clients implemented a targeted semantic backdoor that attempts to misclassify images of red cars as frogs. Assuming that there are $k$ arbitrary Byzantine clients, we replaced a proportion of each training batch in their respective training data with images of red cars, mislabelled as frogs. We term this model $M_{bk}^i$ and the initial global model $M_g^i$. These clients perform additional epochs of training (six for Byzantine clients, compared to two for non-Byzantine clients), using stochastic gradient descent with a momentum value of 0.9 and a learning rate of 0.01. They also scale their final model update by submitting $M_{sk}^i = M_g^i + \gamma \times (M_{bk}^i - M_g^i)$ as the final model for aggregation. For the purposes of our investigation, we set the scaling factor to $\gamma = 0.33$, as we found that setting $\gamma = \frac{n}{f\eta}$ to achieve total model replacement resulted in the Byzantine models being excluded from aggregation by all robust aggregation techniques tested. Each client received an equally-sized shard of the training dataset.

To measure the backdoor's effect on the global model, we tracked two metrics for each round of training:

    \begin{enumerate}[leftmargin=*]
        \item \emph{Accuracy}: the sparse categorical accuracy taken across the validation set for the entire CIFAR-10 dataset, consisting of 10,000 out of 60,000 total images labelled with one of ten categories.
        \item \emph{Misclassification}: the sparse categorical accuracy taken across the validation set of the backdoored items, consisting of 8,168 images of red cars, mislabelled as frogs. These images of red cars were generated from 1,021 images of red cars from the CIFAR-10 dataset. Each image was flipped horizontally, then their saturation and brightness were randomly adjusted.
    \end{enumerate}

For this analysis, we compared Simeon against the same aggregation methods as before: coordinate-wise median, Krum, Bulyan (with Krum as the selection function), and federated averaging. We measured the accuracy and misclassification of the global model (as described above) after each round of training and aggregation, also running the experiment run for 250 rounds of training. As before, Krum and Bulyan were provided an accurate upper bound for the number of Byzantine clients. Simeon was configured with $\epsilon = 10^{-7}$. Again, we also tested the performance of Krum and Bulyan when given an inaccurate upper bound by setting $f = 0$ for Krum and $f = 2$ for Bulyan in the 20\% and 30\% Byzantine test cases. To serve as an additional point of comparison, results for a control example were collected, involving an identical set-up, using federated averaging as the aggregator, but with zero Byzantine clients.

\paragraph{Results and analysis}

\begin{figure}
        \begin{center}
            \includegraphics[width=.7\linewidth]{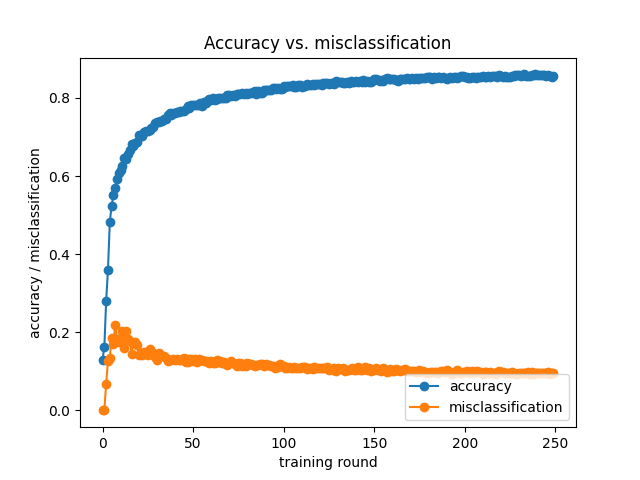}
        \end{center}
        \vspace{-1.2em}
        \caption[Backdoor attack control]{Accuracy and rate of misclassification with federated averaging and \emph{zero} Byzantine clients over 250 rounds of training.}
    \end{figure}

    \begin{figure*}
        \begin{center}
            \includegraphics[width=.33\linewidth]{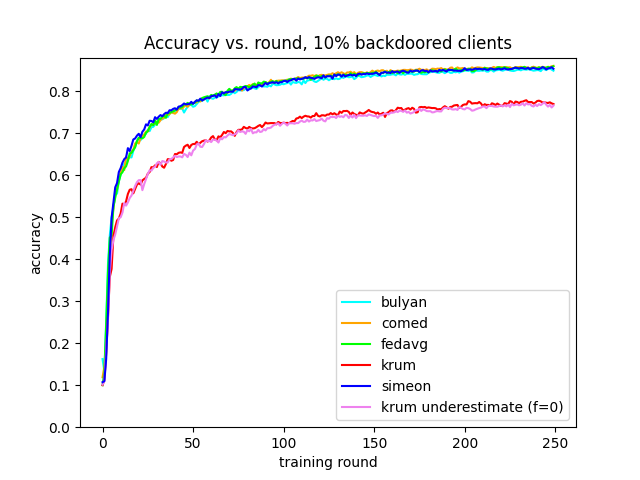}
            \quad \includegraphics[width=.33\linewidth]{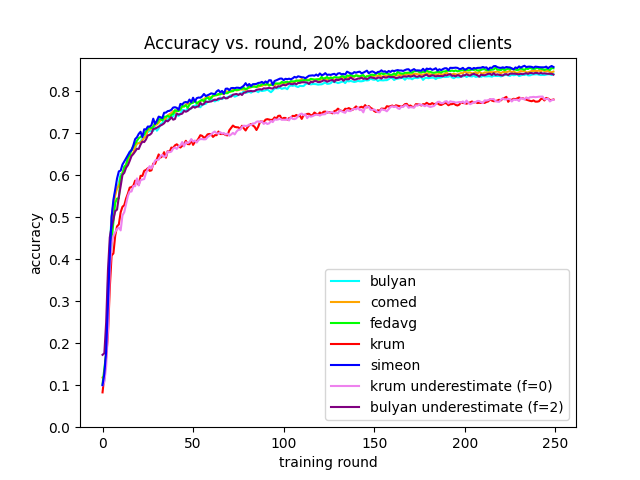}
            \quad \includegraphics[width=.33\linewidth]{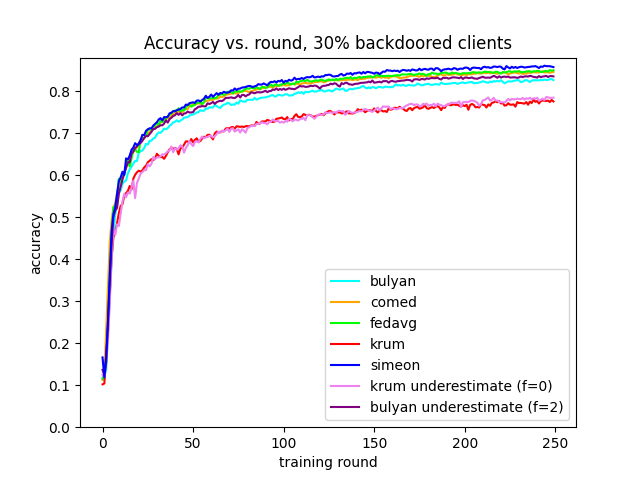}
        \end{center}
        \vspace{-1.2em}
        \caption[Backdoor attack accuracy]{Left to Right: accuracy when the backdoor attack is performed, with a  scaling value of $\gamma = 0.33$} 
    \end{figure*}
    
    \begin{figure*}
        \begin{center}
            \includegraphics[width=.33\linewidth]{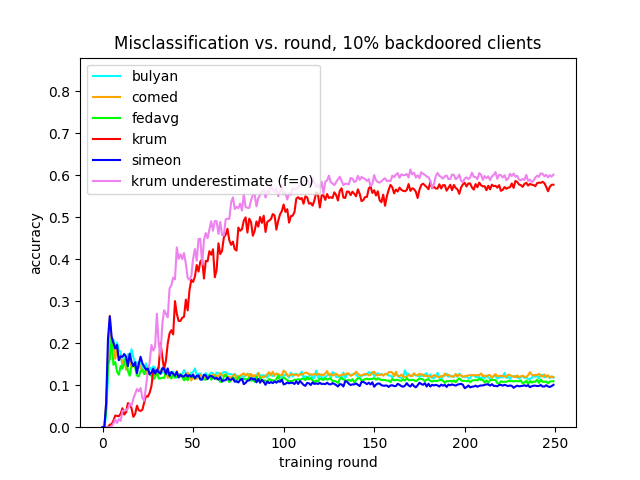}
            \quad \includegraphics[width=.33\linewidth]{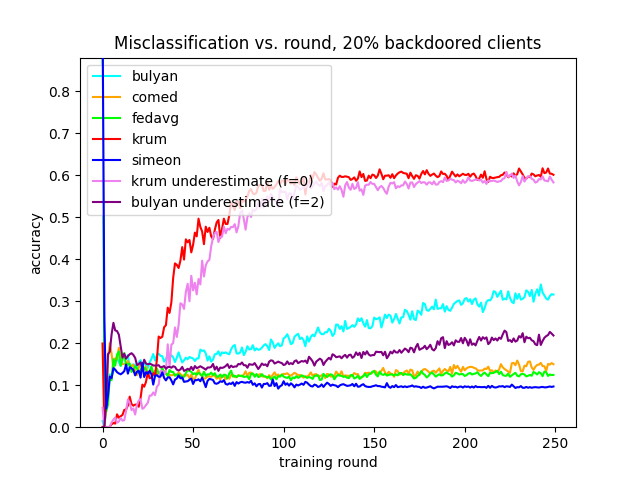}
            \quad \includegraphics[width=.33\linewidth]{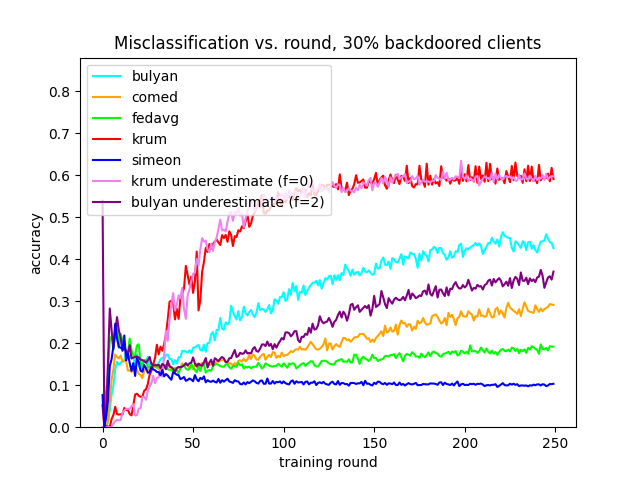}
        \end{center}
        \vspace{-1.2em}
        \caption[Backdoor attack misclassification]{Left to Right: rate of misclassification when the backdoor attack is performed, with a scaling value of $\gamma = 0.33$} 
    \end{figure*}

The results are shown in Figure~\ref{fig:XXXXXXX}.
We note several common patterns in the results. The output of each global model after the first iteration is effectively random initially, causing drastic differences between the rate of misclassification between different algorithms during the early training rounds.As training progresses, the rate of misclassification initially rises, then falls, then rises again. This behaviour was observed for the single-shot attack in \cite{bagdasaryan2019}. We note an upward trajectory to the misclassification of all algorithms, except for Simeon. We also note an effect on the overall accuracy of final model produced by the backdoor-affected algorithms: we speculate that this is because of the presence of other red images in the data set, which could be classified incorrectly due to the influence of the backdoor.

Consistent with observations in \cite{bagdasaryan2019}, we see that Krum displays clear vulnerability in all three test cases \cite{bagdasaryan2019}. This is because the selection of a single backdoored client will cause backdoored clients to be more likely to be selected in the future, due to the lowered scaling value, coupled with the increased number of training epochs for backdoored clients, which results in a model that is relatively close to non-Byzantine models. The use of the Euclidean distance as the scoring mechanism accounts for this behaviour, as the high dimensionality of the model space renders the Euclidean distance ineffective at detecting anomalous models when only a small number of weights are different from the consensus \cite{elmhamdi2018}. 

We note that Bulyan is also vulnerable to this backdoor when a larger proportion of malicious clients are present. This is because Bulyan relies on Krum for selecting models for aggregation: since Krum selects compromised models
for aggregation, they are present in the subsequent coordinate-wise trimmed mean step. We believe Bulyan exhibits vulnerability to this attack because setting $\gamma <$ 1.0 decreases the variance between the Byzantine and non-Byzantine models. This results in compromised models influencing the final model output after the trimmed mean step. The use of Krum sampling creates a self-reinforcing effect, as models closer to the current global model will be preferred during initial sampling during subsequent rounds. We speculate that the final averaging step weakens the effect of the backdoor because the individual contributions of each backdoored client are `watered-down' by the presence of the non-backdoored models. We note that the underestimate case actually performs better than the base case, because the increased number of selected clients results in a greater weakening effect due to the final averaging.

We find that coordinate-wise median is somewhat vulnerable to this attack. Whilst the median value for most coordinates is unlikely to be close to the compromised value, we hypothesise that the lowered scaling value brings the compromised models closer to the median for each coordinate. Despite this, this leads to lower effectiveness than other algorithms, but still a noticeable upward trend to the rate of misclassification.

Federated averaging performs unexpectedly well for this attack, likely due to the lowered scaling value, which does not leverage the possibility of total model replacement. The decreased scaling, coupling with the averaging effect reduces the effectiveness of the backdoor.

Simeon, we find, is robust to this attack. Simeon does not use the Euclidean distance between models to inform aggregation, and instead measures the variance of each model from the current reputation vector. A maximum likelihood estimation technique is used to evaluate the models against one another to inform the credibility score, and hence, the final estimate. Benign clients are unlikely to agree with Byzantine ones, and hence Byzantine clients will receive low credibility scores, thereby reducing their impact on the aggregated value. By using the variance rather than the Euclidean distance to score models, we find that Simeon is more sensitive to divergences between models.

\section{Sybil attack}
\label{sec:sybil}
\paragraph{Experiment implementation and design}
We examined the effect of a sybil attack on the four algorithms described above, involving performing an initial 29 rounds of training with 20 clients, of which two implemented the backdoor attack as described above. At the beginning of the 30th round of training, ten additional Byzantine clients were introduced to the system, all of which also implemented the backdoor attack, with the same configuration parameters as in the previous experiment. 250 total rounds of training were performed. After the injection of the new clients, there were 30 clients in total, 12 of which were Byzantine. All clients trained using an equally-sized shard of the CIFAR-10 training dataset. These shards were made smaller after the injection of the new clients to ensure that the entire dataset is used, and that no clients receive overlapping data items.

As before, we measured the \emph{accuracy} and \emph{misclassification}, using the same method described above. After the injection of the Byzantine clients, neither Simeon, Krum nor Bulyan were reconfigured to account for the changed upper bound of Byzantine clients, to simulate the rapid influx of an unexpectedly large number of malicious learning agents. Hence, we set $f = 2$ for Krum and Bulyan, and $\epsilon = 10^{-7}$ for Simeon, and did not change either parameter throughout training.

\paragraph{Results and analysis}
The results are shown in Figure~\ref{fig:XXXXXXX}.
As before, we find that federated averaging is vulnerable and fails to train after the injection of the additional Byzantine clients. We also find that Krum remains robust with our original value for $\gamma$, and is similarly vulnerable when $\gamma = 0.33$. In the latter case, the sybil attack amplified the strength of the backdoor after being introduced at the start of round 30.
    \begin{figure}
        \begin{center}
        \includegraphics[width=.6\linewidth]{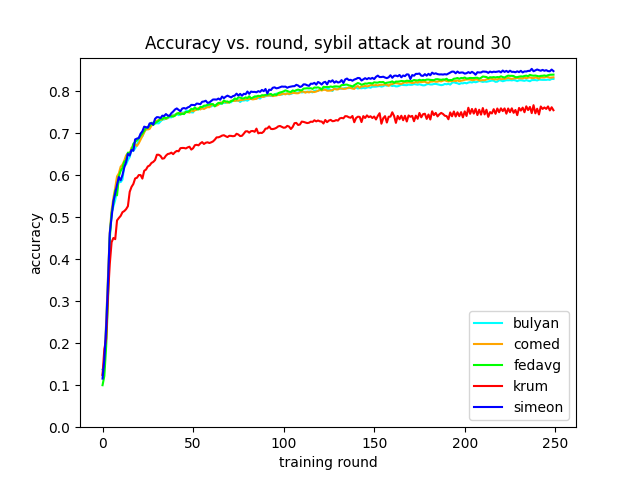}
        \includegraphics[width=.6\linewidth]{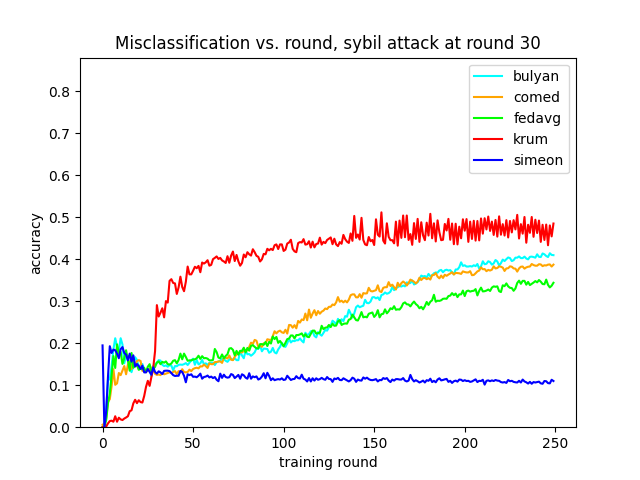}
        \end{center}
        \vspace{-1.2em}
        \caption[Sybil attack misclassification]{Rate of accuracy and misclassification respectively when the backdoor attack is performed, with a scaling value of $\gamma = 0.33$ over 250 rounds of training. An additional 10 clients were introduced at round 30 to implement a sybil attack}
    \end{figure}

We note that coordinate-wise median is vulnerable after the injection of the sybil attack at round 30: the marked increased in the number of Byzantine clients allows the median of each coordinate to be shifted more strongly towards those of the malicious models. Assuming that the differences between the benign models for each coordinate and the global model are randomly-distributed, only a single benign model needs to submit a value close to the maliciously-chosen value for the backdoored value to be chosen for that coordinate. In other words, the increased density of backdoored clients will skew the median for each coordinate towards the backdoored value.

This attack, when applied against Bulyan, achieves an elevated level of misclassification in the presence of an overwhelming number of sybil clients. After the clients are injected at the start of round 30, we note that the level of misclassification increases and continues to increase, similarly to Krum. Recalling that Bulyan provides strong robustness guarantees when $f <= \frac{n-3}{4}$ \cite{elmhamdi2018}, it is clear that when $f = 12$ and $n = 30$ that this upper bound is exceeded. Beyond this threshold, Krum will select backdoored models for aggregation, and the subsequent trimmed mean step will not exclude all of the backdoored clients, permitting them to affect the aggregated value.

Simeon also exhibits robustness to this attack, enabled by the use of a precision parameter, $\epsilon$ to control the number of iterations performed. Prior to round 30, Simeon converges within 4 iterations per round, whilst from round 30 onwards, the algorithm converges within 15 iterations per round. This is because the newly-introduced clients introduce a greater degree of variance to the initial estimate. Recalling that Simeon iterates until ${\rm RMSE}(E, E_k) < \epsilon$, it is natural to see that a greater degree of variance between the input models results in a larger number of iterations performed. Upon convergence from round 30 onwards, the 12 Byzantine models were generally weighted in total less than 6\% in the final model, despite forming 40\% of the number of clients in the system. We attribute Simeon's robustness against this attack involving drastically changing numbers of clients to this mechanism, and identify Simeon's robustness against sybil attacks as a key advantage of our approach.


\section{Increasing scaling attack}
\label{sec:increasing}
\paragraph{Experiment implementation and design}
To ascertain the effect of different scaling values on Simeon, we examined a situation where $\gamma$ varies over time. Using the same network architecture, optimiser, loss function, and attack method as described previously, an experiment was run where $\gamma$ was uniformly increased from $\gamma = 0$ to $\gamma = 0.66$ from the first round of training to round 150, then kept steady until a total of 250 rounds of training were completed. Out of the 20 clients, two exhibited Byzantine behaviour by implementing the semantic backdoor as described earlier.

\paragraph{Results and analysis}

    \begin{figure}
        \begin{center}
        \includegraphics[width=.6\linewidth]{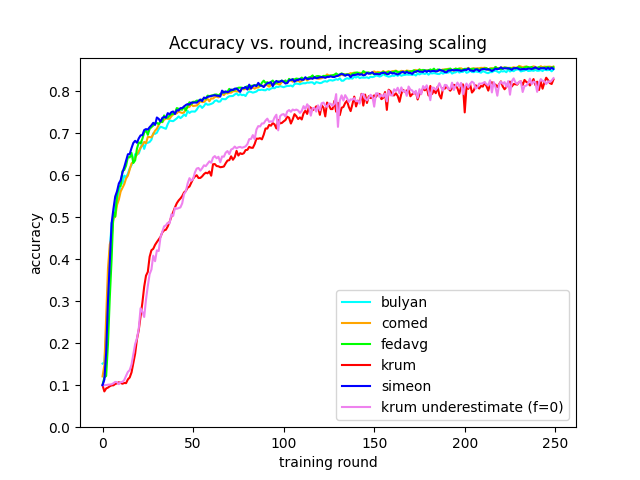}
        \includegraphics[width=.6\linewidth]{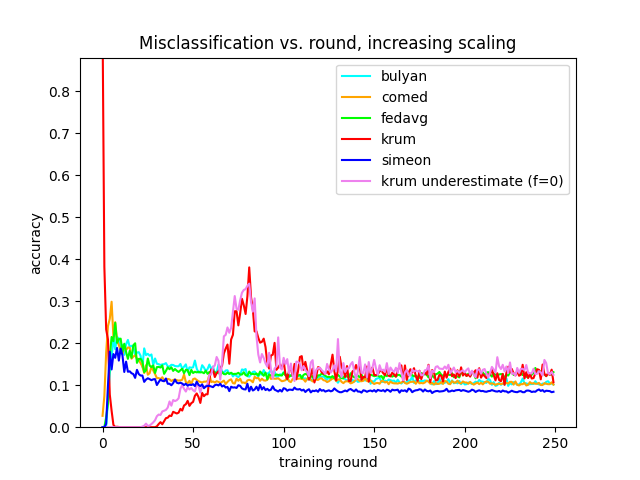}
        \end{center}
        \vspace{-1.2em}
        \caption[Increasing scaling attack misclassification]{Rate of accuracy and misclassification respectively when the backdoor attack is performed, with a  scaling value that increases linearly from $\gamma = 0.0$ to $\gamma = 0.66$ at round 150. A total of 250 rounds of training were performed.}
    \end{figure}
The results are shown in Figure~\ref{fig:XXXXXXX}.
We note that all algorithms other than Krum exhibit a rising-then-falling rate of misclassification: we hypothesise that this pattern occurs because at low levels of accuracy, there is an elevated chance of incorrectly classifying any given image as the attacker-selected category. As training continues and the model's accuracy rises, the chance of misclassification is reduced. We found that this pattern was present even in a control example involving no malicious clients.

Krum initially exhibits near-100\% misclassification, followed by a low misclassification, before peaking at 38\% misclassification, then falling to a final misclassification of 10\%. The initial near-100\% misclassification is due to the model initially outputting the same class for almost all images, giving an extremely low accuracy rating of 9\%. The peak of 38\% misclassification occurs at round 81 when $\gamma = 0.3564$. We believe this value corresponds to a `sweet spot' where the Euclidean distance between compromised models and benign models is low enough to allow Krum to select them, but the effect of the scaling is such that the global model remains in a region where the backdoor is effective. As $\gamma$ increases, the distance between the backdoored models and the benign models grows, resulting in the exclusion of the backdoored models from selection, causing the rate of misclassification to fall as the backdoor is `forgotten'.

Bulyan does not display evidence of vulnerability. The reason for this is twofold: firstly, because the Byzantine clients only comprise 10\% of the overall number of clients present, the final averaging step greatly reduces the influence of the backdoored weights, even if they are not excluded from aggregation. This phenomenon is observable in the previous test case where 10\% of clients were backdoored and $\gamma$ was not increased. Secondly, as $\gamma$ is increased, Krum becomes less likely to select backdoored models for subsequent aggregation because the Euclidean distance between the backdoored models and the benign models also increases. This effect of $\gamma$ is accentuated further compared to the original investigation \cite{bagdasaryan2019} due to the reduced number of parameters in our model. Ultimately, this causes Bulyan to exhibit no significant elevation in the rate of misclassification.

Coordinate-wise median also does not display evidence of vulnerability. This is because the proportion of Byzantine clients is not large enough to significantly affect the median. Moreover, as $\gamma$ rises, the Byzantine models are shifted away from the median, resulting in their exclusion from aggregation.

We also found that, as before, federated averaging does not display evidence of elevated misclassification. This is due to the reduced scaling value, which does not leverage the possibility of achieving total model replacement. Ultimately, we found federated averaging to be unaffected by this attack, if only because the scaling value was kept below 1.0.

By iteratively refining estimates of the variance of each model rather than relying on the pairwise Euclidean distance between models, Simeon maintains resilience to this attack. When the scaling value is low, the variance between the backdoored models and the benign models is also low -- in these cases, the malicious models are weighted roughly equally with the non-malicious models, but because of the very low scaling value the backdoor is ineffective. As the scaling value rises, the variance between the backdoored and non-backdoored models also rise, resulting in lower credibility scores for the malicious models. By round 70, when $\gamma = 0.3036$, the two backdoored clients received weightings totalling less than 1\% of the final global model. Given this, it is clear that Simeon cannot be induced to give high weightings to malicious models using a scaling-based attacks such as this.

\section{Conclusion}
Our investigation has demonstrated that Simeon is resilient to rudimentary model poisoning attacks as well as more sophisticated attacks with proven effectiveness on current-best algorithms. In particular, we determined that Simeon is resilient to a targeted poisoning attack involving scaling that is known to affect Krum and other Byzantine-robust aggregation algorithms \cite{bagdasaryan2019}.

Additionally, we found that Simeon is able to minimise the impact that sybils have on aggregation, even when they almost form a majority of clients. In contrast, Bulyan \cite{elmhamdi2018} and Krum \cite{blanchard2017} exhibited vulnerability to sybil attacks because they require prior configuration with an upper bound, and do not guarantee robustness beyond that upper bound. The literature has identified that the need for prior knowledge of the maximum density of attackers is a key vulnerability of current state-of-the-art algorithms \cite{fung2019}: Simeon’s $\epsilon$ parameter enables more or fewer iterations to be performed as the degree of variance between models changes, without requiring the number of attackers to be known. We believe this is a key advantage of Simeon compared to other approaches.



\bibliography{icml2021ref}

\begin{thebibliography}{19}
\providecommand{\natexlab}[1]{#1}
\providecommand{\url}[1]{\texttt{#1}}
\expandafter\ifx\csname urlstyle\endcsname\relax
  \providecommand{\doi}[1]{doi: #1}\else
  \providecommand{\doi}{doi: \begingroup \urlstyle{rm}\Url}\fi

\bibitem[Bagdasaryan et~al.(2020)Bagdasaryan, Veit, Hua, Estrin, and
  Shmatikov]{bagdasaryan2019}
Bagdasaryan, E., Veit, A., Hua, Y., Estrin, D., and Shmatikov, V.
\newblock How to backdoor federated learning.
\newblock In Chiappa, S. and Calandra, R. (eds.), \emph{Proceedings of the
  Twenty Third International Conference on Artificial Intelligence and
  Statistics}, volume 108 of \emph{Proceedings of Machine Learning Research},
  pp.\  2938--2948. PMLR, 26--28 Aug 2020.
\newblock URL \url{http://proceedings.mlr.press/v108/bagdasaryan20a.html}.

\bibitem[Baruch et~al.(2019)Baruch, Baruch, and Goldberg]{baruch2019}
Baruch, G., Baruch, M., and Goldberg, Y.
\newblock A little is enough: Circumventing defenses for distributed learning.
\newblock In Wallach, H., Larochelle, H., Beygelzimer, A., d~Alch\'{e}-Buc, F.,
  Fox, E., and Garnett, R. (eds.), \emph{Advances in Neural Information
  Processing Systems (NeurIPS)}, volume~32, pp.\  8635--8645. Curran
  Associates, Inc., 2019.
\newblock URL
  \url{https://proceedings.neurips.cc/paper/2019/file/ec1c59141046cd1866bbbcdfb6ae31d4-Paper.pdf}.

\bibitem[Beaufays et~al.(2019)Beaufays, Rao, Mathews, and
  Ramaswamy]{beaufays2019}
Beaufays, F., Rao, K., Mathews, R., and Ramaswamy, S.
\newblock Federated learning for emoji prediction in a mobile keyboard, 2019.
\newblock URL \url{https://arxiv.org/abs/1906.04329}.

\bibitem[Bhagoji et~al.(2019)Bhagoji, Chakraborty, Mittal, and
  Calo]{bhagoji2019}
Bhagoji, A.~N., Chakraborty, S., Mittal, P., and Calo, S.
\newblock Analyzing federated learning through an adversarial lens.
\newblock In Chaudhuri, K. and Salakhutdinov, R. (eds.), \emph{Proceedings of
  the 36th International Conference on Machine Learning}, volume~97 of
  \emph{Proceedings of Machine Learning Research}, pp.\  634--643. PMLR, 09--15
  Jun 2019.
\newblock URL \url{http://proceedings.mlr.press/v97/bhagoji19a.html}.

\bibitem[Blanchard et~al.(2017)Blanchard, El~Mhamdi, Guerraoui, and
  Stainer]{blanchard2017}
Blanchard, P., El~Mhamdi, E.~M., Guerraoui, R., and Stainer, J.
\newblock Brief announcement: Byzantine-tolerant machine learning.
\newblock In \emph{Proceedings of the ACM Symposium on Principles of
  Distributed Computing}, PODC '17, pp.\  455–457, New York, NY, USA, 2017.
  Association for Computing Machinery.
\newblock ISBN 9781450349925.
\newblock \doi{10.1145/3087801.3087861}.
\newblock URL \url{https://doi.org/10.1145/3087801.3087861}.

\bibitem[de~Kerchove \& Van~Dooren(2010)de~Kerchove and Van~Dooren]{dekerchove}
de~Kerchove, C. and Van~Dooren, P.
\newblock Iterative filtering in reputation systems.
\newblock \emph{SIAM Journal of Matrix Analysis and Applications}, 31\penalty0
  (4):\penalty0 1812–1834, March 2010.
\newblock ISSN 0895-4798.
\newblock \doi{10.1137/090748196}.
\newblock URL \url{https://doi.org/10.1137/090748196}.

\bibitem[El~Mhamdi et~al.(2018)El~Mhamdi, Guerraoui, and Rouault]{elmhamdi2018}
El~Mhamdi, E.~M., Guerraoui, R., and Rouault, S.
\newblock The hidden vulnerability of distributed learning in {B}yzantium.
\newblock In Dy, J. and Krause, A. (eds.), \emph{Proceedings of the 35th
  International Conference on Machine Learning}, volume~80 of \emph{Proceedings
  of Machine Learning Research}, pp.\  3521--3530, Stockholm Sweden, 10--15 Jul
  2018. PMLR.
\newblock URL \url{http://proceedings.mlr.press/v80/mhamdi18a.html}.

\bibitem[Fang et~al.(2020)Fang, Cao, Jia, and Gong]{fang2019}
Fang, M., Cao, X., Jia, J., and Gong, N.~Z.
\newblock Local model poisoning attacks to byzantine-robust federated learning.
\newblock In Capkun, S. and Roesner, F. (eds.), \emph{29th {USENIX} Security
  Symposium, {USENIX} Security 2020, August 12-14, 2020}, pp.\  1605--1622.
  {USENIX} Association, 2020.
\newblock URL
  \url{https://www.usenix.org/conference/usenixsecurity20/presentation/fang}.

\bibitem[Fung et~al.(2020)Fung, Yoon, and Beschastnikh]{fung2019}
Fung, C., Yoon, C. J.~M., and Beschastnikh, I.
\newblock The limitations of federated learning in sybil settings.
\newblock In \emph{23rd International Symposium on Research in Attacks,
  Intrusions and Defenses ({RAID} 2020)}, pp.\  301--316, San Sebastian,
  October 2020. {USENIX} Association.
\newblock ISBN 978-1-939133-18-2.
\newblock URL
  \url{https://www.usenix.org/conference/raid2020/presentation/fung}.

\bibitem[Grama et~al.(2020)Grama, Musat, Muñoz-González, Passerat-Palmbach,
  Rueckert, and Alansary]{grama2020robust}
Grama, M., Musat, M., Muñoz-González, L., Passerat-Palmbach, J., Rueckert,
  D., and Alansary, A.
\newblock Robust aggregation for adaptive privacy preserving federated learning
  in healthcare, 2020.
\newblock URL \url{https://arxiv.org/abs/2009.08294}.

\bibitem[Hard et~al.(2018)Hard, Kiddon, Ramage, Beaufays, Eichner, Rao,
  Mathews, and Augenstein]{hard2018}
Hard, A., Kiddon, C.~M., Ramage, D., Beaufays, F., Eichner, H., Rao, K.,
  Mathews, R., and Augenstein, S.
\newblock Federated learning for mobile keyboard prediction, 2018.
\newblock URL \url{https://arxiv.org/abs/1811.03604}.

\bibitem[Kairouz \& McMahan(2021)Kairouz and McMahan]{kairouz2019advances}
Kairouz, P. and McMahan, H.~B.
\newblock Advances and open problems in federated learning.
\newblock \emph{Foundations and Trends in Machine Learning}, 14\penalty0
  (1):\penalty0 --, 2021.
\newblock ISSN 1935-8237.
\newblock \doi{10.1561/2200000083}.
\newblock URL \url{http://dx.doi.org/10.1561/2200000083}.

\bibitem[Lamport et~al.(1982)Lamport, Shostak, and Pease]{lamport1982}
Lamport, L., Shostak, R., and Pease, M.
\newblock The byzantine generals problem.
\newblock \emph{ACM Transactions on Programming Languages and Systems},
  4\penalty0 (3):\penalty0 382–401, July 1982.
\newblock ISSN 0164-0925.
\newblock \doi{10.1145/357172.357176}.
\newblock URL \url{https://doi.org/10.1145/357172.357176}.

\bibitem[{Laureti, P.} et~al.(2006){Laureti, P.}, {Moret, L.}, {Zhang, Y.-C.},
  and {Yu, Y.-K.}]{laureti}
{Laureti, P.}, {Moret, L.}, {Zhang, Y.-C.}, and {Yu, Y.-K.}
\newblock Information filtering via iterative refinement.
\newblock \emph{EPL (Europhysics Letters)}, 75\penalty0 (6):\penalty0
  1006--1012, 2006.
\newblock \doi{10.1209/epl/i2006-10204-8}.
\newblock URL \url{https://doi.org/10.1209/epl/i2006-10204-8}.

\bibitem[Li et~al.(2019)Li, Wen, Wu, Hu, Wang, and He]{li2019}
Li, Q., Wen, Z., Wu, Z., Hu, S., Wang, N., and He, B.
\newblock A survey on federated learning systems: Vision, hype and reality for
  data privacy and protection, 2019.
\newblock URL \url{https://arxiv.org/abs/1907.09693}.

\bibitem[Lyu et~al.(2020)Lyu, Yu, Zhao, and Yang]{lyu2020threats}
Lyu, L., Yu, H., Zhao, J., and Yang, Q.
\newblock \emph{Threats to Federated Learning}, pp.\  3--16.
\newblock Springer International Publishing, Cham, 2020.
\newblock ISBN 978-3-030-63076-8.
\newblock \doi{10.1007/978-3-030-63076-8_1}.
\newblock URL \url{https://doi.org/10.1007/978-3-030-63076-8_1}.

\bibitem[McMahan et~al.(2017)McMahan, Moore, Ramage, Hampson, and
  y~Arcas]{mcmahan2017}
McMahan, B., Moore, E., Ramage, D., Hampson, S., and y~Arcas, B.~A.
\newblock Communication-efficient learning of deep networks from decentralized
  data.
\newblock In Singh, A. and Zhu, X.~J. (eds.), \emph{Proceedings of the 20th
  International Conference on Artificial Intelligence and Statistics, {AISTATS}
  2017, 20-22 April 2017, Fort Lauderdale, FL, {USA}}, volume~54 of
  \emph{Proceedings of Machine Learning Research}, pp.\  1273--1282. {PMLR},
  2017.
\newblock URL \url{http://proceedings.mlr.press/v54/mcmahan17a.html}.

\bibitem[Rezvani et~al.(2013)Rezvani, Ignjatovic, Bertino, and
  Jha]{rezvani2014}
Rezvani, M., Ignjatovic, A., Bertino, E., and Jha, S.
\newblock A robust iterative filtering technique for wireless sensor networks
  in the presence of malicious attacks.
\newblock In \emph{Proceedings of the 11th ACM Conference on Embedded Networked
  Sensor Systems}, SenSys '13, New York, NY, USA, 2013. Association for
  Computing Machinery.
\newblock ISBN 9781450320276.
\newblock \doi{10.1145/2517351.2517394}.
\newblock URL \url{https://doi.org/10.1145/2517351.2517394}.

\bibitem[Yin et~al.(2018)Yin, Chen, Kannan, and Bartlett]{yin2018}
Yin, D., Chen, Y., Kannan, R., and Bartlett, P.
\newblock {B}yzantine-robust distributed learning: Towards optimal statistical
  rates.
\newblock In Dy, J. and Krause, A. (eds.), \emph{Proceedings of the 35th
  International Conference on Machine Learning}, volume~80 of \emph{Proceedings
  of Machine Learning Research}, pp.\  5650--5659, Stockholm Sweden, 10--15 Jul
  2018. PMLR.
\newblock URL \url{http://proceedings.mlr.press/v80/yin18a.html}.

\end{thebibliography}
\bibliographystyle{icml2021}

\end{document}